\documentclass[runningheads]{llncs}

\usepackage[T1]{fontenc}
\usepackage[utf8]{inputenc}
\usepackage{newtxtext}
\usepackage[varvw]{newtxmath}

\usepackage{graphicx}
\usepackage{soul}
\usepackage{textcomp}
\usepackage{xcolor}
\usepackage{multirow}
\usepackage{enumerate}

\usepackage{algorithm}
\usepackage[noend]{algpseudocode}

\usepackage{etoolbox}

\usepackage[hyphens]{url}
\usepackage[hidelinks]{hyperref}

\graphicspath{{./imgs/}}
\algrenewcommand\alglinenumber[1]{\scriptsize #1:}
\algrenewcommand\algorithmicindent{1.0em}

\algnewcommand\algorithmicforeach{\textbf{for each}}
\algdef{S}[FOR]{ForEach}[1]
    {\algorithmicforeach\ #1\ \algorithmicdo}

\begin{document}
\title{LocalIntent: Outcome-Specific Structured Generation for Backend-Compatible Smart Building Automation with Local LLMs} 
\author{Tianyang Liu\inst{1} \and
Jiayue Wang\inst{1} \and
Zhanhao Cheng\inst{1} \and Changyang Wang\inst{1} \and Shaoshu Zhu\inst{2} \and Zhenghao Wu\inst{2} \and Hadi Tabatabaee~Malazi\inst{1,}\inst{2}}
\institute{Beijing Dublin International College, University College Dublin, Beijing, China \\
\and
School of Computer Science, University College Dublin, Dublin, Ireland\\
\email {%\{Tianyang.Liu, Jiayue.Wang, Zhanhao.Cheng, Changyang.Wang, Shaoshu.Zhu1, Zhenghao.Wu\}@ucdconnect.ie}, 
Hadi.TabatabaeeMalazi@ucd.ie}}
\titlerunning{LocalIntent: Outcome-Specific Structured Generation}
\authorrunning{Tianyang Liu et al.}
\maketitle              % typeset the header of the contribution
\begin{abstract}
Locally deployed large language models (LLMs) can reduce reliance on remote cloud services for natural language smart building automation, but backend-facing processing requires correct outcome selection, alignment with local capabilities, and valid structured outputs. We investigate whether a complete outcome-specific prompting design improves compact-model outputs beyond generic route-then-generate prompting. The proposed \emph{LocalIntent} approach routes requests to ordinary or persistent-rule branches: the former produces commands, clarifications, or rejections, while the latter generates trigger--action rules, followed by deterministic validation. We evaluate four locally executed Gemma and Qwen models spanning 2B--9B parameters on 101 non-persistent and 54 persistent-rule records, comparing \emph{Monolithic} prompting, generic two-stage prompting, LocalIntent, and validation-guided repair. Across the evaluated datasets, \emph{LocalIntent} achieves higher intent understanding, device identification, and action/parameter correctness than generic two-stage prompting, while clarification/mode correctness remains model-dependent. The largest gains occur for rules, where \emph{LocalIntent} improves intent understanding from 26.4\% to 80.6\%, device identification from 59.3\% to 84.7\%, rule-mode correctness from 49.1\% to 98.6\%, and action/parameter correctness from 5.6\% to 44.9\%. For non-persistent requests, pooled gains are smaller: intent understanding increases from 39.6\% to 53.2\%, and action/parameter correctness from 30.7\% to 42.8\%.
\end{abstract}

\section{Introduction}
Smart buildings coordinate lighting, ventilation, blinds, and sensor-driven services. Natural language interfaces can improve access, but requests require different backend outcomes: \emph{"turn on the meeting-room lights"} is an immediate command, whereas \emph{"turn on the meeting-room lights whenever occupancy is detected"} defines a persistent rule. Incomplete requests require clarification; unsupported requests require rejection.

Intent processing must select the requested behaviour and backend outcome, then generate a representation the platform can process. \emph{Automation-output correctness} requires correct outcome selection, schema compliance, grounding in available devices and sensors, supported actions and parameters, and semantic consistency. A well-formed output is insufficient if it invents capabilities, guesses missing information, or executes future-triggered behaviour immediately.

These requirements are particularly important for local LLMs, which must process requests near the managed environment while relying on local capability information and limited computational resources. On-device LLMs have supported Home Assistant intent detection and response generation~\cite{birkmose2025ondevice} and privacy-oriented personalised assistance~\cite{huang2026homellama}. However, local processing does not ensure backend-ready output: compact models must still interpret temporal scope, select an outcome, resolve entities, and generate structured output within resource and response-time constraints.

Prior work examines LLM-based device control~\cite{king2024sasha}, IoT program synthesis~\cite{shen2025autoiot,shen2025gpiot}, rule authoring~\cite{gallo2024rulebotpp}, and structured generation~\cite{lu2025learning,wang2025slot}. Evaluations also expose failures on invalid, ambiguous, and multi-device requests~\cite{li2025homebench}. These strands usually treat intent interpretation, program construction, rule authoring, or schema compliance separately. They provide limited evidence on the complete backend-facing transformation from a request to a command, stored rule, clarification, or rejection, followed by a representation valid for that outcome. Existing approaches rarely evaluate whether outcome selection itself should determine the generation constraints and validation criteria used for backend-facing automation. Our hypothesis is that the complete LocalIntent design achieves higher automation-output correctness than generic route-then-generate prompting, particularly for persistent rules. 

We introduce \emph{LocalIntent}, which routes requests to an ordinary or persistent-rule branch. The rule branch generates a predicate-bearing trigger and action. The ordinary branch produces a target--action--parameter command, clarification, or rejection under non-execution guardrails. A deterministic validator tests structure, registered entities, supported actions and parameters, and mechanically testable rule consistency. We compare \emph{LocalIntent} with \emph{Monolithic} prompting and \emph{TwoStageGeneric}, a generic route-then-generate design. The methods differ in routing labels, prompts, schemas, context presentation, and fallbacks; thus, we compare complete designs without isolating individual mechanisms. We also assess validation-guided repair and local inference latency. \emph{LocalIntent}'s clearest differences occur on persistent rules; non-persistent differences are smaller and model-dependent, and repair adds little aggregate benefit.

This paper makes four contributions: 
\begin{enumerate}
    \item We formulate local LLM-based smart building automation as structured generation with four outcomes: immediate commands, persistent rules, clarification, and rejection, each with distinct temporal semantics and validity conditions.
    \item We introduce \emph{LocalIntent} with ordinary/rule routing, command and non-execution guardrails, persistent-rule mapping, and automatic local-capability validation.
    \item We construct two controlled evaluation datasets: 101 non-persistent requests covering commands, clarifications, and rejections, and 54 persistent-rule requests. We evaluate four local Gemma and Qwen models (2B--9B parameters).
    \item We release implementation and evaluation artefacts in a public repository\footnote{\url{https://github.com/ucd-soc2/localIntent}}.
\end{enumerate}

\section{Related Work}
\label{sec:related-work}
Natural-language smart-device interfaces may initiate physical actions, making ambiguous or unsupported requests consequential. Sasha~\cite{king2024sasha} translates underspecified goals into device plans but reports unsuitable and hallucinated behaviour. HomeBench~\cite{li2025homebench} exposes difficulties with valid and invalid, single- and multi-device instructions, while Calò et al.~\cite{calo2024enhancing} use multimodal information to resolve ambiguity. For local deployment, these safety concerns coexist with privacy and resource constraints. At the infrastructure level, distributed service placement and workload orchestration address response-time and backhaul-traffic optimisation in multi-access edge computing~\cite{9592440}, while sustainable serverless edge scheduling distinguishes energy efficiency from carbon awareness~\cite{10.1145/3773274.3774929}. Birkmose et al.~\cite{birkmose2025ondevice} demonstrate on-device Home Assistant intent detection, slot filling, and JSON calls. HomeLLaMA~\cite{huang2026homellama} explores personalised assistance, and CIDER~\cite{jeong2025cider} combines small models with an ontology. AIoT Smart Home~\cite{rivkin2025aiot} studies autonomous device-control agents, while SmartIntent~\cite{sun2025smartintent} integrates compact models into serverless building automation. \emph{LocalIntent} instead examines backend-facing selection and validity across commands, persistent rules, clarifications, and rejections.

AutoIoT~\cite{shen2025autoiot} and GPIoT~\cite{shen2025gpiot} investigate IoT program synthesis. Almond~\cite{campagna2017almond} maps language to executable ThingTalk programs. RuleBot++~\cite{gallo2024rulebotpp} creates and modifies Home Assistant rules, and Ply~\cite{10.1145/3746059.3747638} supports inspection and user control in trigger--action programming. Everyday-automation studies highlight challenges in object/service selection, timing, and expressiveness~\cite{10.1145/3734863}. These approaches primarily address program or rule construction rather than the preceding choice between execution, rule creation, clarification, and rejection.

Decomposed Prompting~\cite{khot2023decomposed} separates complex tasks into stages. Schema Reinforcement Learning~\cite{lu2025learning}, SLOT~\cite{wang2025slot}, and format comparisons~\cite{yuan2026structuredoutput} address structured-output compliance, while constrained decoding~\cite{beurerkellner2024guiding} can enforce formal languages at potential efficiency or accuracy costs. Yet a valid structure need not imply correct outcome selection or support for generated entities, actions, and parameters. LLM-based parameter collection for energy optimisation~\cite{michelon2025hems} and feedback-based refinement~\cite{madaan2023selfrefine} address related issues. \emph{LocalIntent} integrates outcome selection, branch-specific generation, capability validation, and optional repair. We evaluate complete designs, not isolated mechanisms.

\section{LocalIntent System Architecture}
\label{sec:architecture} 
\begin{figure}[h]
\centering
\includegraphics[width=0.8\textwidth]{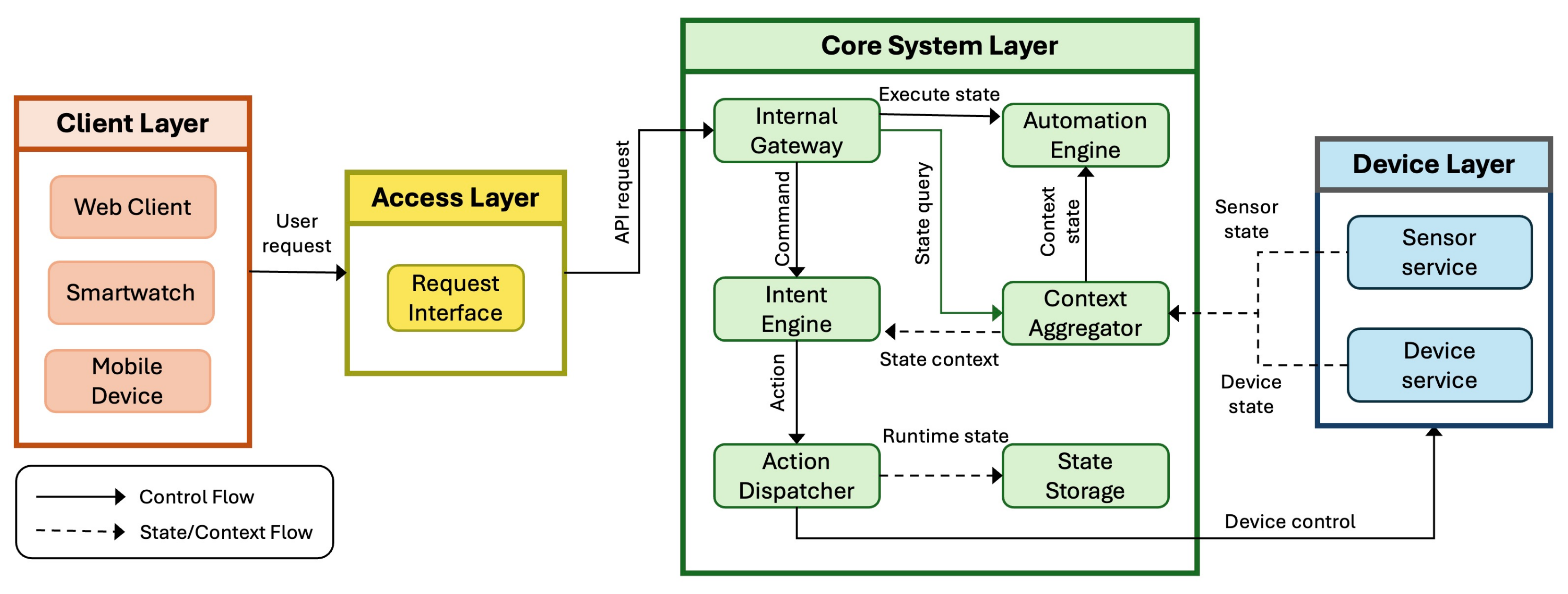}
\caption{\emph{LocalIntent} within a layered smart building automation platform.}
\label{fig:system_architecture}
\end{figure}

The system comprises three layers: access receives requests; the core manages context, routing, generation, and validation; the device layer exposes device and sensor services (Figure~\ref{fig:system_architecture}). The prototype implements the Internal Gateway, Intent Engine, Automation Engine, Context Aggregator, and State Storage. The Action Dispatcher and Automation Engine are intended to receive validated commands and rules. Experiments assess outputs and latency without physical actuation or concurrent state management.

%The system comprises three layers: access receives requests; the core maintains context and handles routing, generation, and validation; the device layer exposes device and sensor services (Figure~\ref{fig:system_architecture}). The prototype implements the Internal Gateway, Intent Engine, Automation Engine, Context Aggregator, and State Storage. The Action Dispatcher and Automation Engine are intended destinations for validated commands and rules. Experiments assess outputs and latency without physical actuation or concurrent state management.

Formally, for request $u$ at time $t$, let $ C_t=\langle Z,D,S,X_t,\mathcal{K}\rangle$ denote the local context. $Z$, $D$, and $S$ are configured location zones, actionable devices, and observable sensors. $X_t$ is the record's state snapshot, and registry $\mathcal{K}$ associates entities with identifiers, types, zones, and supported capabilities. For devices, it records actions and parameter definitions, including types, units, ranges, or admissible values. For sensors, it records observable attributes and supported trigger predicates. $D$ and $S$ may overlap. 

Let $\mathcal{P}= \{\textsc{Command},\textsc{Rule},
\textsc{Clarification},\textsc{Rejection}\} $
be the outcomes. Each outcome has specification $\Sigma_p=\langle\mathcal{S}_p,\Phi_p\rangle$, where $\mathcal{S}_p$ is its parser-accepted structure and $\Phi_p$ its semantic constraints. A \textsc{Command} is an immediate target--action--parameter operation; a \textsc{Rule} is a persistent trigger--action structure; \textsc{Clarification} requests information needed for a supported completion; and \textsc{Rejection} denotes functionality for which no supported completion exists within the registered capabilities or platform scope. \emph{LocalIntent} routes over the coarser set $\mathcal{H}=\{\textsc{Ordinary},\textsc{Rule}\}$, where \textsc{Ordinary} produces a command, clarification, or rejection. With mapping $\pi(\textsc{Rule})=\textsc{Rule}$ and $\pi(p)=\textsc{Ordinary}$ otherwise, the branch specification is the collection of admissible outcome alternatives, $\Sigma_h=\{\Sigma_p:p\in\mathcal P,\ \pi(p)=h\}$.
Thus, the Rule branch contains only the Rule outcome specification, whereas the Ordinary branch contains the Command, Clarification, and Rejection specifications.

\emph{LocalIntent} separates routing, generation, and validation:
\[
h=\mathcal{R}(u,C_t), h\in\mathcal{H}, r=\mathcal{G}_{h}(u,C_t), (\hat y,e_p)=\operatorname{ParseNorm}(r), 
\mathcal{V}_{h}(\hat{y},C_t,\Sigma_h)\rightarrow(b,e).
\]
$\mathcal{R}$ selects a branch, $\mathcal{G}_{h}$ produces raw response $r$, and $\operatorname{ParseNorm}$ parses and normalises it into candidate $\hat y$ or parse diagnostics $e_p$. Deterministic validator $\mathcal{V}_{h}$ then applies the parsing-independent structural, capability, parameter, and rule-consistency checks applicable to the generated outcome. $b\in\{0,1\}$ indicates success and $e$ contains structured diagnostics. Validated commands and rules are intended for the Action Dispatcher and Automation Engine, respectively; clarifications and rejections return via the access layer. 

The \emph{LocalIntent+SemanticRepair} variant permits at most one additional LLM call after validation failure. Its prompt contains the request, context, raw and normalised outputs, validation diagnostics, and diagnostic-specific guidance. The candidate must pass the same validator and a reference-free minimal-change guard. The guard preserves the outcome family and device/action identities unless the diagnostic permits reconstruction. Missing-parameter failures may complete required fields while preserving the normalised semantic payload. The candidate replaces the initial output only if it passes both checks. Otherwise, the initial output is retained only for paired experimental scoring and remains ineligible for downstream hand-off if validator-rejected. 

\section{Outcome-Specific Structured Generation}
\label{sec:prompt}
\emph{LocalIntent} uses one local LLM call for ordinary/rule routing and another for branch generation. ``Outcome-specific'' denotes the distinct output alternatives and validity conditions embedded within these two branch prompts, not four separate generation calls. For compactness, $\mathcal S_h$ and $\Phi_h$ below denote the structures and constraints induced by the outcome specifications collected in $\Sigma_h$. Their mechanically evaluable subset is checked after generation.

\subsection{Outcome Routing}
For a request $u$ and automation context $C_t$, the routing call predicts $\hat{h}=\mathcal{R}_{\theta}(u,C_t)$, $\hat{h}\in\{\textsc{Ordinary},\textsc{Rule}\}$, where $\mathcal{R}_{\theta}$ is the routing call to model $\theta$.

If the \emph{LocalIntent} gate does not return a valid label, a predefined temporal-cue heuristic selects \textsc{Ordinary} or \textsc{Rule}. This heuristic recovers only a temporal branch label. It does not independently re-establish completeness, capability support, or representability and can therefore violate the normal Rule-eligibility policy. 

Reference outcomes follow the policy:
\begin{small}
\[
p^\ast=
\begin{cases}
\textsc{Rejection}, &
\text{if the requested functionality has no supported completion},\\
\textsc{Clarification}, &
\text{if a supported completion exists but required information is unresolved},\\
\textsc{Rule}, &
\text{if the request is complete, supported, persistent, and representable},\\
\textsc{Command}, &
\text{if the request is complete, supported, and immediate}.
\end{cases}
\]
\end{small}

The reference branch is $h^\ast=\pi(p^\ast)$. Accordingly, branch routing is correct when $\hat h=h^\ast$, although branch-routing accuracy is not reported as a separate aggregate metric.

A supported completion uses registered entities, actions, and valid parameters and is representable in the appropriate output schema without changing the requested functionality. Missing information therefore leads to \textsc{Clarification} only when such a completion remains possible. Unavailable functionality leads to \textsc{Rejection}. Temporal cues such as \emph{"when"}, \emph{"whenever"}, and \emph{"if"} may indicate persistence, but rule selection also requires a representable trigger and supported action. An imperative with an unresolved target requires clarification, not execution. 

\subsection{Outcome-Specific Generation} 
After routing, \emph{LocalIntent} assembles a prompt for the predicted branch: $q_{\hat{h}}=
q_{\mathrm{shared}}
\oplus u
\oplus \operatorname{ctx}_{\hat{h}}(C_t)
\oplus \mathcal{S}_{\hat{h}}
\oplus \Phi_{\hat{h}},$
where $\oplus$ denotes deterministic assembly, $q_{\mathrm{shared}}$ defines the task, $\operatorname{ctx}_{\hat{h}}$ serialises branch-relevant context, $\mathcal{S}_{\hat{h}}$ defines structure, and $\Phi_{\hat{h}}$ supplies constraints. 

The two-branch design separates eligible persistent rules from ordinary requests, which retain commands, clarifications, and rejections. This delays the ordinary outcome decision until generation can apply capability and completeness constraints, while restricting rule generation to trigger--action structures. Such conditioning aims to reduce incompatible outputs but does not guarantee correct routing or semantic accuracy.

The ordinary branch receives zones, devices, supported actions, parameter definitions, relevant states, and the sensor/predicate capability information needed to assess potentially persistent requests. It produces commands, clarifications, or rejections within their respective guardrails. The rule branch receives the full trigger-relevant sensor and predicate context and must produce exactly one trigger and one action. Capability information enables the ordinary branch to distinguish unresolved from unsupported persistent requests, while the rule branch uses it to ground the trigger and action in registered capabilities. For a parseable output, the generated backend outcome is
\begin{small}
\[
\hat p=\begin{cases}
\textsc{Rule},&\hat h=\textsc{Rule},\\
\operatorname{Outcome}(\hat y),&\hat h=\textsc{Ordinary}, 
\end{cases}
\]
\end{small}

where $\operatorname{Outcome}(\hat y)$ reads the command, clarification, or rejection discriminator from the normalised Ordinary-branch output. 

\begin{table}[!b]
\centering
\scriptsize
\caption{Outcome-specific fields and generation constraints.}
\label{tab:output-schemas}
\setlength{\tabcolsep}{3.2pt}
\renewcommand{\arraystretch}{1.08}
\begin{tabular}{p{0.12\linewidth} p{0.20\linewidth} p{0.60\linewidth}}
\hline
\textbf{Outcome} &
\textbf{Semantic fields} &
\textbf{Generation guidance} \\
\hline
\textsc{Command} &
target, action, parameters &
Preserve immediate semantics. Select a registered target and supported action. Provide the required valid parameters. Do not introduce rule fields or infer unresolved information. \\
\hline
\textsc{Rule} &
predicate-bearing trigger, action &
Preserve persistent semantics. Map the trigger to a registered sensor and supported predicate. Select a supported action. Maintain consistency between the trigger predicate and action. Do not represent the action as immediate execution. \\
\hline
\textsc{Clarification} &
unresolved element, question &
Identify the information preventing an otherwise supported completion. Request only the information needed to continue. Do not generate command or rule fields. \\
\hline
\textsc{Rejection} &
unsupported element, reason &
Identify the requested functionality that is unavailable under the registered capabilities or platform scope. Do not generate command or rule fields. \\
\hline
\end{tabular}
\end{table}

In the rule schema, a trigger contains a sensor, comparison operator, and threshold. Its action contains a target, supported action, and required parameters. Clarification and rejection are explicitly non-executable: the former requests information that could yield a valid output, while the latter identifies functionality unavailable under current capabilities. Neither includes a command or rule payload. Table~\ref{tab:output-schemas} groups parser-accepted fields by semantic role rather than reproducing literal implementation field names. 

\subsection{Validation Interface} 
The deterministic validator checks structure, registered entities, supported actions, parameter domains, prohibited fields, and mechanically testable rule consistency. Whether an output correctly expresses the user's goal is assessed separately against reference annotations. Validation failures prevent downstream hand-off (Algorithm~\ref{alg:localintent}).
\begin{algorithm}[ht]
\scriptsize
\caption{Branch-specific structured generation in \emph{LocalIntent}}
\label{alg:localintent}
\begin{algorithmic}[1]
\Require Request $u$, local context $C_t$
\Ensure Outcome $\hat p$ and candidate $\hat{y}$ when parseable, status $b$, diagnostics $e$
\State $\hat{h}\gets\mathcal{R}_{\theta}(u,C_t)$
\If{$\hat{h}\notin\{\textsc{Ordinary},\textsc{Rule}\}$}
    \State $\hat{h}\gets\operatorname{TemporalFallback}(u)$
\EndIf
\State Construct $q_{\hat{h}}$ from $u$, branch-relevant context,
$\mathcal{S}_{\hat{h}}$, and $\Phi_{\hat{h}}$
\State $r\gets\operatorname{LLM}_{\theta}(q_{\hat{h}})$
\State $(\hat{y},e_p)\gets\operatorname{ParseNorm}(r)$
\If{$e_p\neq\varnothing$}
    \State $\hat p\gets\varnothing$; $\hat y\gets\varnothing$; $b\gets0$; $e\gets e_p$
\Else
    \State $\hat p\gets\textsc{Rule}$ if $\hat h=\textsc{Rule}$; otherwise $\hat p\gets\operatorname{Outcome}(\hat y)$
    \State $(b,e)\gets
    \mathcal{V}_{\hat{h}}(\hat{y},C_t,\Sigma_{\hat{h}})$
\EndIf
\If{$b=0$}
    \State Mark the request as not ready for downstream hand-off
\EndIf
\State \Return $(\hat p,\hat y,b,e)$
\end{algorithmic}
\end{algorithm}

\section{Performance Evaluation}
\label{sec:performance}
We compare \emph{LocalIntent} with \emph{Monolithic} and \emph{TwoStageGeneric} using six dataset-specific diagnostic metrics, paired repair analysis, and local inference latency.

\emph{LocalIntent} and \emph{TwoStageGeneric} share models, requests, capability snapshots, parsing, validation, and scoring, but differ in routing, prompts, output organisation, context presentation, and fallbacks. \emph{TwoStageGeneric} receives $\mathcal{S}_p$ without LocalIntent-specific routing guidance or semantic constraints $\Phi_p$. Invalid \emph{TwoStageGeneric} routes default to \textsc{Clarification}, whereas \emph{LocalIntent} uses a temporal-cue fallback. Thus, comparisons assess complete designs rather than individual mechanisms. 

\subsection{Evaluation Design}
\textit{RQ1} compares \emph{Monolithic}, \emph{TwoStageGeneric}, and \emph{LocalIntent} on reference-based correctness. \textit{RQ2} examines differences between non-persistent requests and rules, while \textit{RQ3} evaluates local inference latency. \emph{Monolithic} selects and generates an outcome in one call; \emph{TwoStageGeneric} performs four-way routing, and \emph{LocalIntent} uses ordinary/rule routing. Paired initial and final outputs assess \emph{LocalIntent+SemanticRepair}. Generation settings are shared across methods for each model.

\subsection{Datasets and Local Models}
Each JSONL record contains an instruction, device and sensor state, timestamp, and reference output. The non-persistent dataset has 101 requests: 54 commands, 31 clarifications, and 16 rejection cases. Its references use \texttt{needsClarification}, \texttt{message}, and \texttt{result}. Rule references contain one predicate-bearing trigger and one action within a \texttt{rule} object. The rule dataset contains 54 records representing 49 distinct instruction--state pairs. Five pairs have two reference actions subsequently judged admissible. All 54 records are retained, giving these five pairs double weight. Reference-based metrics do not exhaustively establish semantic validity. The dataset covers six trigger-sensor and five action-device types using \texttt{>=} or \texttt{<=}, state-derived thresholds, supported actions, and legal parameters. It contains only clear executable rules and does not test clarification or rejection of rule requests.

Written specifications defined task boundaries, capabilities, value ranges, schemas, and consistency constraints. Candidate records were manually checked for task type, state consistency, logical validity, and JSON syntax. LLM-assisted checks were advisory, and reported issues were manually adjudicated before correction or removal. Because prompt exemplars overlap some benchmark formulations, results measure controlled prompt-development performance rather than held-out generalisation.

Outputs were generated with Ollama~0.32.1 on an Apple M4 MacBook Air with 16\,GB memory using \texttt{gemma2:2b}, \texttt{qwen2.5:3b}, \texttt{gemma3:4b}, and \texttt{gemma2:9b}. We used a 4096-token context window and a 120\,s timeout. Temperature, top-$p$, top-$k$, and seed were not explicitly set and followed Ollama or model defaults. Models remained loaded in memory during evaluation. 

\subsection{Metrics}
Each normalised output receives six binary, dataset-specific diagnostic scores. For non-persistent requests, intent understanding (IU) requires an exact device--action match. For rules, it requires the reference trigger sensor, action device, and action, but not operator, value, or parameters. Device identification (DI) matches action-device sets and excludes rule-trigger sensors. Clarification handling (CH) matches \texttt{needsClarification} for non-persistent requests and rule/non-rule mode for rules. Action/parameter correctness (APC) requires matching actions and parameters, with tolerances of $\pm1$ for temperature, brightness, level, and channel and $\pm5$ for volume and \texttt{openPercentage}. Categorical values require exact agreement. Rule APC also requires an exact trigger object. Output-format compliance (OFC) checks normalised schema, registered identifiers, supported actions, and parameter keys but not agreement with the reference outcome. Language consistency (LC) checks message language for non-persistent outputs but only the top-level rule structure. Clarification/rejection message quality is unmeasured.

Except for one \emph{LocalIntent+SemanticRepair} output for which no metric record exists, missing or unparsable outputs receive zero on all metrics. Rates are mean binary scores over records within each dataset. Metrics overlap, and CH and LC change meaning across datasets, so results are stratified and the metrics are not interpreted as independent dimensions. Repair is assessed through paired metric changes; Figure~\ref{fig:stage-latency} reports repair-call counts. Latency uses observed mean, median, and 95th percentile. 

\subsection{Baseline Comparison}
\begin{figure}[t]
    \centering
    \includegraphics[width=0.92\linewidth]{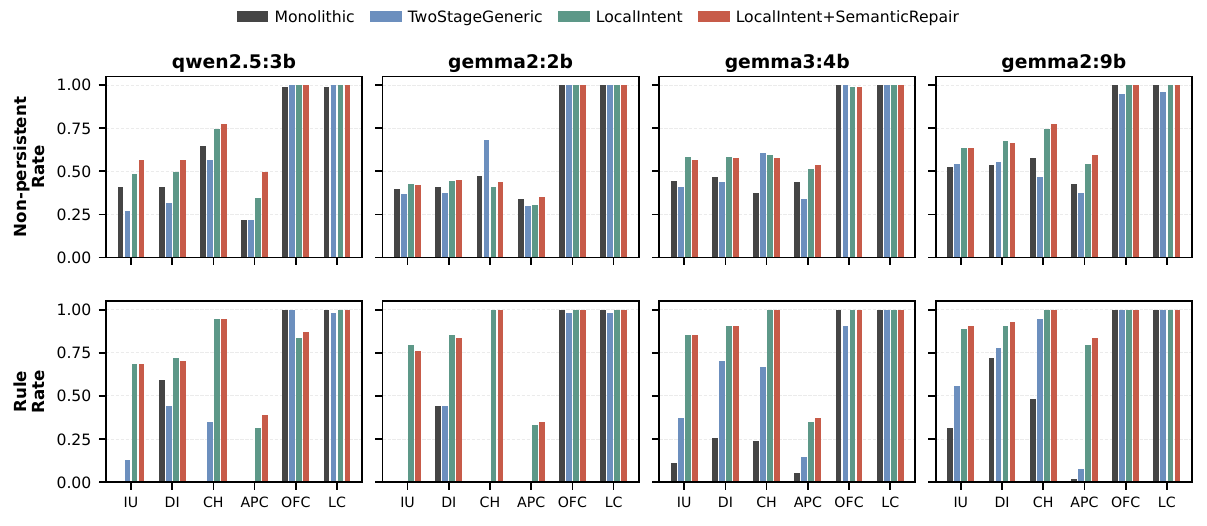}
    \caption{Six dataset-specific metrics by method and model.}    
    \label{fig:method-comparison}
\end{figure}

\emph{LocalIntent} records higher pooled IU, DI, CH, and APC than \emph{TwoStageGeneric} on both datasets, with larger gains on rules. RQ1--RQ2 concern the three primary methods; Figure~\ref{fig:method-comparison} presents all methods across four Qwen and Gemma models and both datasets.

On non-persistent requests, pooled IU, DI, CH, and APC are 0.396, 0.421, 0.579, and 0.307 for \emph{TwoStageGeneric}, and 0.532, 0.550, 0.621, and 0.428 for \emph{LocalIntent}, each across 404 model--record evaluations. \emph{LocalIntent+SemanticRepair} obtains 0.546, 0.563, 0.640, and 0.494 across 403 available evaluations because one output lacks a metric record. Both \emph{LocalIntent} conditions exceed \emph{TwoStageGeneric} on all four pooled metrics; the repair condition also has higher pooled rates than the base condition. Both conditions exceed \emph{TwoStageGeneric} in IU, DI, and APC for all four models. For \emph{LocalIntent}, CH is higher for two models and lower for two; the repair condition exceeds \emph{TwoStageGeneric} in CH for two models. OFC and LC remain near the ceiling in most non-persistent conditions, offering less discrimination than the semantic metrics.

On rules, pooled IU, DI, CH, and APC are 0.264, 0.593, 0.491, and 0.056 for \emph{TwoStageGeneric}; 0.806, 0.847, 0.986, and 0.449 for \emph{LocalIntent}; and 0.801, 0.843, 0.986, and 0.486 for \emph{LocalIntent+SemanticRepair}, each across 216 model--record evaluations. \emph{LocalIntent} exceeds \emph{TwoStageGeneric} on all four metrics for every model. Compared descriptively with the separate \emph{LocalIntent} condition, the repair condition has slightly lower pooled IU and DI, identical CH, and higher APC. Most gains over \emph{TwoStageGeneric} are therefore already present in base \emph{LocalIntent}.

In the separately generated \emph{LocalIntent} condition, 511 of 620 outputs passed deterministic validation, but 277 of those failed at least one reference-based metric. Validator acceptance therefore did not establish full reference agreement.

\subsection{Validation-Guided Repair}
\label{sec:repair}
\begin{figure}[!t]
    \centering
    \includegraphics[width=0.95\linewidth]{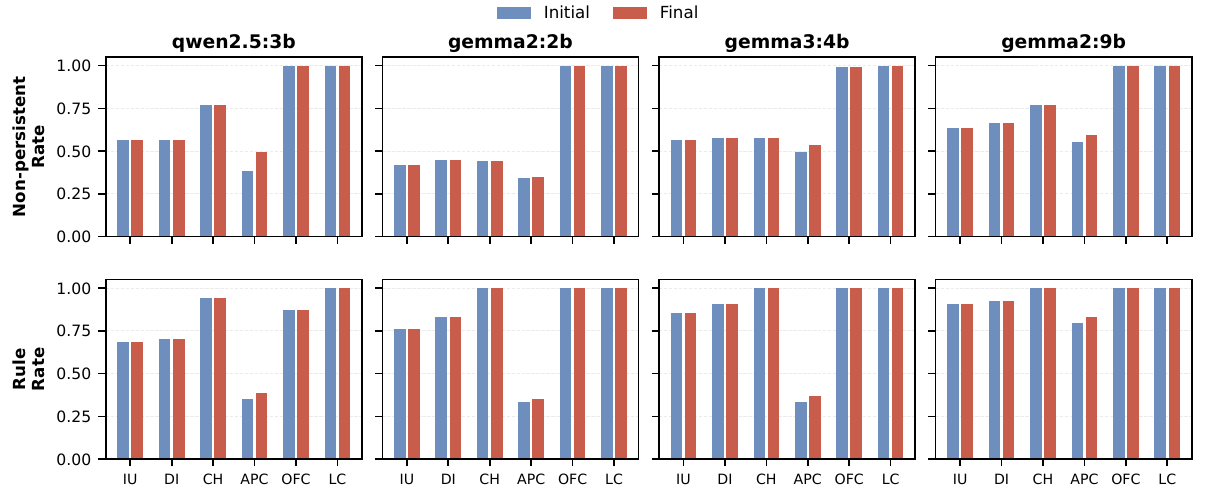}
    \caption{Paired initial/final repair metrics; rejected repairs retain initial outputs.} 
    \label{fig:repair-comparison}
\end{figure}

Figure~\ref{fig:repair-comparison} compares paired initial (blue) and selected final (red) outputs under \emph{LocalIntent+SemanticRepair}. Across all eight model--dataset combinations, final APC is no lower than initial APC, while paired IU, DI, CH, OFC, and LC totals remain unchanged. Net gains in APC-correct evaluations are +13 for \texttt{qwen2.5:3b}, +2 for \texttt{gemma2:2b}, +6 for \texttt{gemma3:4b}, and +6 for \texttt{gemma2:9b}, totalling 27. Most gains occur on non-persistent requests; rules show smaller positive gains for all four models. 

\subsection{Local Inference Latency}
\label{sec:latency} 
Observed local inference latency varies with model, request type, and summary statistic across \emph{LocalIntent}, \emph{TwoStageGeneric}, and the repair variant.

\begin{figure}[!b]
    \centering
    % Top-left: P95
    \begin{minipage}[t]{0.499\linewidth}
        \vspace{0pt}
        \centering
        \includegraphics[width=\linewidth]{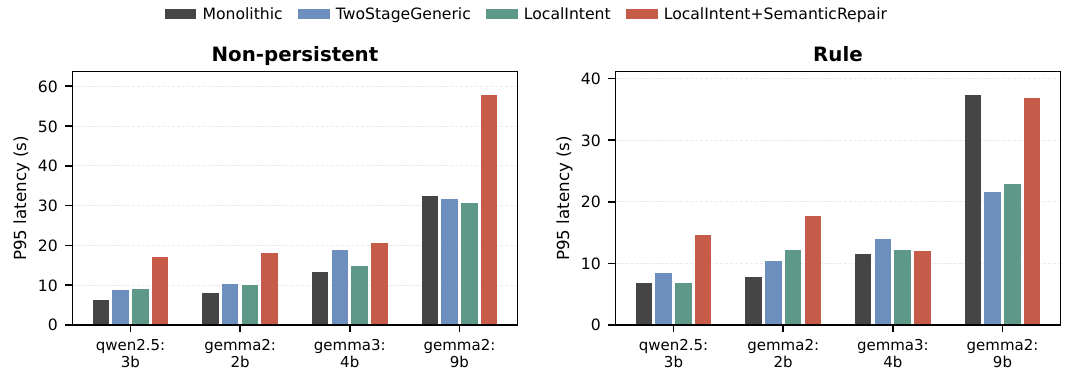}
        \caption{Observed end-to-end P95 latency by method, model, and dataset.}
        \label{fig:local-latency-p95}
    \end{minipage}\hfill
    % Top-right: Mean
    \begin{minipage}[t]{0.499\linewidth}
        \vspace{0pt}
        \centering
        \includegraphics[width=\linewidth]{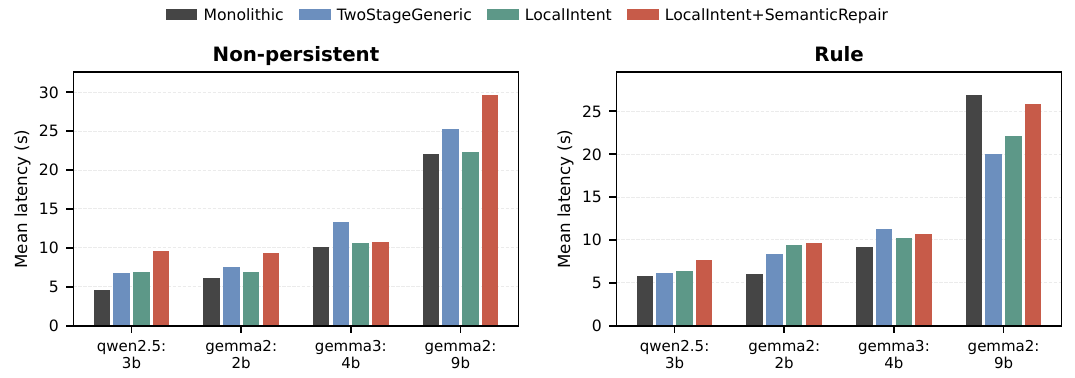}
        \caption{Observed end-to-end mean latency by method, model, and dataset.}
        \label{fig:local-latency-mean}
    \end{minipage}
    \par\medskip
    % Bottom-left: Median
    \begin{minipage}[t]{0.499\linewidth}
        \vspace{0pt}
        \centering
        \includegraphics[width=\linewidth]{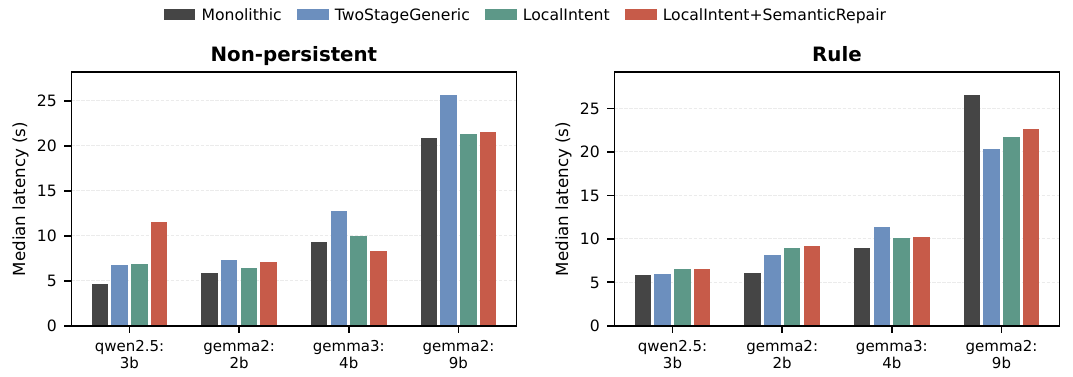}
        \caption{Observed end-to-end median latency by method, model, and dataset.}
        \label{fig:local-latency-median}
    \end{minipage}\hfill
    % Bottom-right: Stages
    \begin{minipage}[t]{0.499\linewidth}
        \vspace{0pt}
        \centering
        \includegraphics[width=\linewidth]{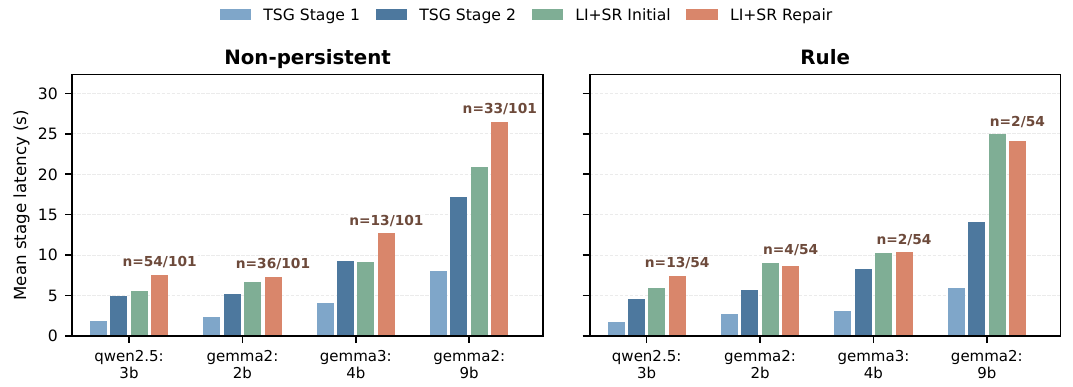}
        \caption{Model-call latency by stage; purposes and counts vary.}
        \label{fig:stage-latency}
    \end{minipage}
\end{figure}

Observed P95 end-to-end latency shows no uniform advantage for either \emph{LocalIntent} or \emph{TwoStageGeneric} (Figure~\ref{fig:local-latency-p95}). On non-persistent requests, \emph{LocalIntent} has lower P95 latency for \texttt{gemma3:4b} and \texttt{gemma2:9b}, slightly lower latency for \texttt{gemma2:2b}, and higher latency for \texttt{qwen2.5:3b}. On rules, it is lower for \texttt{qwen2.5:3b} and \texttt{gemma3:4b}, but higher for \texttt{gemma2:2b} and \texttt{gemma2:9b}. \emph{LocalIntent+SemanticRepair} generally has higher P95 latency than \emph{LocalIntent}, with the largest increase for non-persistent \texttt{gemma2:9b} requests. 

Mean end-to-end latency favours \emph{LocalIntent} over \emph{TwoStageGeneric} for three of four models on non-persistent requests, with the largest reductions for \texttt{gemma3:4b} and \texttt{gemma2:9b} (Figure~\ref{fig:local-latency-mean}). On rules, it is faster only for \texttt{gemma3:4b}. The repair variant has higher mean latency than \emph{LocalIntent} in all eight model--dataset combinations, with the largest increase for non-persistent \texttt{gemma2:9b} requests (29.60 versus 22.26\,s).

Median latency follows similar model-dependent trends to the mean while being less sensitive to long-running requests (Figure~\ref{fig:local-latency-median}). For non-persistent requests, \emph{LocalIntent} has lower medians than \emph{TwoStageGeneric} for all three Gemma models, with nearly identical values for \texttt{qwen2.5:3b}. On rules, \emph{LocalIntent} has a lower median only for \texttt{gemma3:4b}. Repair increases median latency relative to \emph{LocalIntent} in six of eight combinations; the exceptions are non-persistent \texttt{gemma3:4b} and rule-based \texttt{qwen2.5:3b}.

Generation dominates routing latency for \emph{TwoStageGeneric} across models and datasets (Figure~\ref{fig:stage-latency}). For non-persistent \texttt{gemma2:9b} requests, recorded routing and generation times are 8.05 and 17.17\,s, respectively. Under \emph{LocalIntent+SemanticRepair}, initial processing and triggered repair calls are the main measured components. Repair-call latency varies substantially across models and datasets and is particularly high for \texttt{gemma2:9b}, reaching 26.51\,s on non-persistent requests. These measurements help explain the end-to-end trends but remain descriptive rather than controlled component comparisons because stage purposes and invocation counts differ across methods. 

\section{Conclusion}
\label{sec:conclusion}
This paper examined branch-specific structured generation for locally executed smart building automation. \emph{LocalIntent} separates ordinary requests from eligible persistent-rule requests, producing commands, clarifications, or rejections in the ordinary branch and predicate-bearing trigger--action structures in the rule branch. On the evaluated benchmark, the complete design shows its largest correctness gains over generic two-stage prompting for persistent rules, where the baseline frequently fails to produce matching rule-mode outputs. Non-persistent differences are smaller and model-dependent, while paired repair analysis shows targeted action/parameter-correctness (APC) improvements without reductions in the other aggregate metric totals. These descriptive results cover four models, one building configuration, and one generation per method--model--record. Because routing, prompts, schemas, context, and fallbacks change together, component contributions cannot be isolated. Future work should examine held-out multi-building requests, generation stability, targeted ablations, and constrained-decoding baselines. Remaining semantic errors underscore the central lesson: structural and capability validation alone cannot ensure semantic correctness or backend readiness. 

\bibliographystyle{splncs04}
\bibliography{ref}
\end{document}